\documentclass[journal]{IEEEtran}
\usepackage{tabularx}
\usepackage{times}
\usepackage{latexsym}
\usepackage{helvet}
\usepackage{courier}
\usepackage{url}
\usepackage{stfloats}
\usepackage{color}
\usepackage{amsmath}
\usepackage{array}
\usepackage{graphicx}
\usepackage{subfigure}
\usepackage{multirow}
\usepackage{multicol}
\usepackage{CJK}
\usepackage[flushleft]{threeparttable}
\usepackage{float}
\usepackage{mathrsfs}
\usepackage{mathtools}
\usepackage{amsopn}
\usepackage{arydshln} 
\usepackage{bm}
\usepackage{amssymb}
\usepackage{tikz}
\usepackage{pgfplots}
\usepackage{algorithm}
\usepackage{algorithmic}
\usepackage{booktabs}
\usepackage{fontawesome5}

\newcommand\wikiref{\textsc{WikiRef}}
\newcommand\wiki{Wikipedia}

\newcommand\rouge{\textsc{Rouge}}

\newcommand\et{\MakeLowercase{\textit{et al.}}}

\newcommand{\tabincell}[2]{\begin{tabular}{@{}#1@{}}#2\end{tabular}}

\pgfplotsset{compat=1.16}
\def\addlegendimage{\csname pgfplots@addlegendimage\endcsname}

\definecolor{document_color}{HTML}{EFEFEF}
\definecolor{query_color}{HTML}{F5E2E2}
\definecolor{summary_color}{HTML}{C9E7F4}

\ifCLASSINFOpdf
\else
\fi

\ifCLASSOPTIONcompsoc
  \usepackage[nocompress]{cite}
\else
  \usepackage{cite}
\fi
\begin{document}
\begin{CJK*}{UTF8}{gbsn}
%
%
\title{Transforming Wikipedia into Augmented Data \\ for Query-Focused Summarization}
\author{Haichao Zhu, Li Dong, Furu Wei, Bing Qin, Ting Liu
\thanks{

Haichao Zhu, Bing Qin (Corresponding author) and Ting Liu are with the Faculty of Computing, Harbin Institute of Technology, Harbin 150001, China (e-mail: hczhu@ir.hit.edu.cn; qinb@ir.hit.edu.cn; tliu@ir.hit.edu.cn).

Li Dong and Furu Wei are with Microsoft Research Asia, Beijing 100080, China (e-mail: lidong1@microsoft.com; fuwei@microsoft.com).
}
}
\maketitle

\begin{abstract}
  The limited size of existing query-focused summarization datasets renders training data-driven summarization models challenging.
Meanwhile, the manual construction of a query-focused summarization corpus is costly and time-consuming.
In this paper, we use Wikipedia to automatically collect a large query-focused summarization dataset (named \wikiref{}) of more than $280,000$ examples, which can serve as a means of data augmentation.
We also develop a BERT-based query-focused summarization model (Q-BERT) to extract sentences from the documents as summaries.
To better adapt a huge model containing millions of parameters to tiny benchmarks, we identify and fine-tune only a sparse subnetwork, which corresponds to a small fraction of the whole model parameters.
Experimental results on three DUC benchmarks show that the model pre-trained on \wikiref{} has already achieved reasonable performance. After fine-tuning on the specific benchmark datasets, the model with data augmentation outperforms strong comparison systems.
Moreover, both our proposed Q-BERT model and subnetwork fine-tuning further improve the model performance. The dataset is publicly available at \url{https://aka.ms/wikiref.}
\end{abstract}

 
%
\IEEEpeerreviewmaketitle

\section{Introduction}
\label{section1}
\begin{figure}[!t]
\centering
\includegraphics[width=\linewidth]{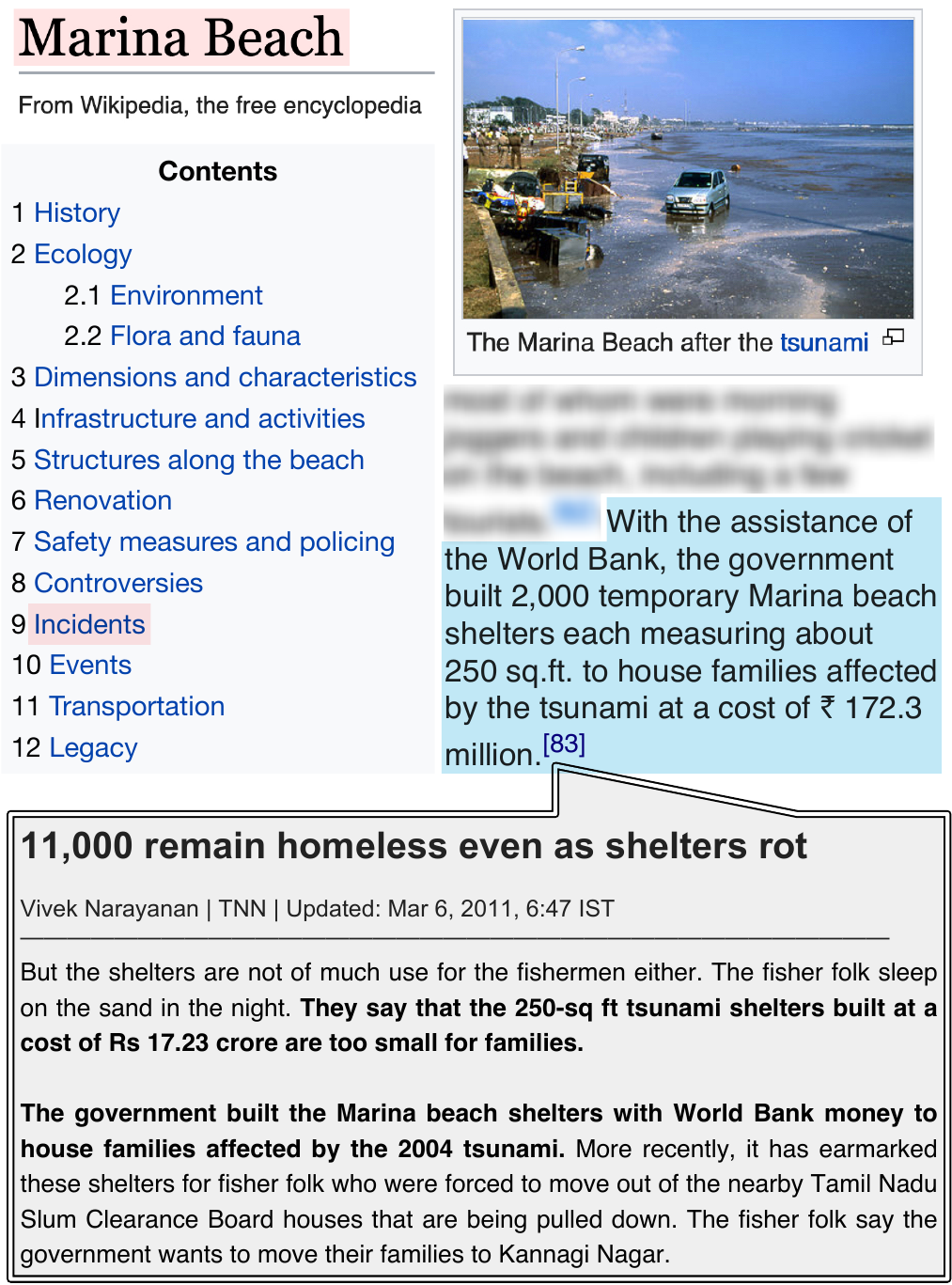}
\caption{Example of automatic query-focused summarization dataset construction.
Given a statement in \wiki{} article ``\textit{Marina Beach}'', we take the body text of citation as the \colorbox{document_color}{document}, use the article title along with section titles (i.e., ``\textit{Marina Beach, Incidents}'') to form a \colorbox{query_color}{query} , and the statement is the \colorbox{summary_color}{summary}.
}
\label{fig:example}
\end{figure}

Query-focused summarization aims to create a brief, well-organized and informative summary for a document with specifications described in the query.
Various unsupervised methods~\cite{mmr,erkan2004lexrank,ilp,qfs_graph, Feigenblat2017,baumel2018query,dual_ces} and supervised methods~\cite{crf,svr,logist_regression,cao-attsum,Ren-crsum,wsl-dl,xu-lapata-2020-coarse} have been proposed for the purpose.
DUC~\cite{duc05overview} 2005~-~2007 are high-quality query-focused summarization benchmarks constructed by humans.
But the limited size renders training neural query-focused summarization models challenging, especially for the data-driven methods.
Meanwhile, the manual construction of a large-scale query-focused summarization dataset is costly and time-consuming.

To advance neural query-focused summarization with limited data, we propose to transform \wiki{} into a large-scale query-focused summarization dataset (named \wikiref{}) as a means of data augmentation.
To automatically construct query-focused summarization examples using \wiki{}, we use the citations of the statements in \wiki{} articles as pivots to align the queries and documents.
Figure~\ref{fig:example} shows an example that is constructed by the proposed method.
We first take the highlighted statement as the summary.
Its supporting citation is expected to provide an adequate context to derive the statement, thus can serve as the source document.
On the other hand, the section titles give a hint about which aspect of the document is the summary's focus. 
Therefore, we use the article title and the section titles of the statement to form the query.
Given that \wiki{} is the largest online encyclopedia, we can automatically construct massive query-focused summarization examples.
At last, we have \wikiref{} dataset of more than $280,000$ examples.

Most extractive summarization models on the DUC benchmarks can be decomposed into two modules, i.e., sentence scoring and sentence selection.
Sentence scoring aims to measure query relevance and sentence salience.
It is well acknowledged that pre-trained language models~\cite{bert, roberta, xlnet, unilm, albert, electra}, e.g., BERT~\cite{bert}, exhibit strong text understanding ability.
In this paper, we develop a BERT-based model (Q-BERT) to score sentences.
The model takes the concatenation of the query and the document as input.
The token-level interactions between the query and the document and their internal interactions are all carried out through multi-head self-attention mechanism~\cite{transformer} in the BERT encoder.
We then apply a query-focused pooling layer on top of the contextual token encodings to get the vector representations of the sentences.
At last, we use a simple linear layer to get the score of each sentence extracted into the summary.
Given sentence scores, we follow the common practice of previous works to select top-ranked sentences with minimal redundancy constraints as the final summary.

Given the proposed extractive model and the large-scale \wikiref{} dataset as augmentation data, we first pre-train the model on the \wikiref{}.
Then we fine-tune the pre-trained model on the benchmark datasets.
The tiny benchmarks only have no more than $100$ examples that can be used to fine-tune the BERT-base model, which contains at least hundreds of millions parameters, e.g., size of BERT-base is $110M$.
Even during fine-tuning, the mismatch between the tiny size of benchmarks and the massive number of parameters poses great challenges to the model optimization.
Therefore, we only fine-tune a small fraction of the whole pre-trained model parameters, which correspond to a sparse subnetwork within the original model.

Experimental results on three DUC benchmarks show that the model achieves competitive performance by fine-tuning, and using \wikiref{} as a means of data augmentation outperforms strong comparsion extractive summarization systems.
The proposed Q-BERT model with pooling layer and the sparse subnetwork fine-tuning (ST) strategy both further improve the model performance.
More importantly, Q-BERT shows that the lacking of large-scale datasets hinders developing more effective data-driven models.
The \wikiref{} is shown to help reveal the effectiveness of these models and is also shown to be an eligible large-scale dataset to advance query-focused summarization research.
Further analysis on augmentation data shows that the data quality is more important than the scale.
Explorations on model structure find that the pooling methods are important to get effective sentence representations.
And the choice of sparse subnetworks for fine-tuning is also critical to the performance besides using fewer parameters.

\begin{figure}[!t]
    \centering
    \includegraphics[width=\columnwidth]{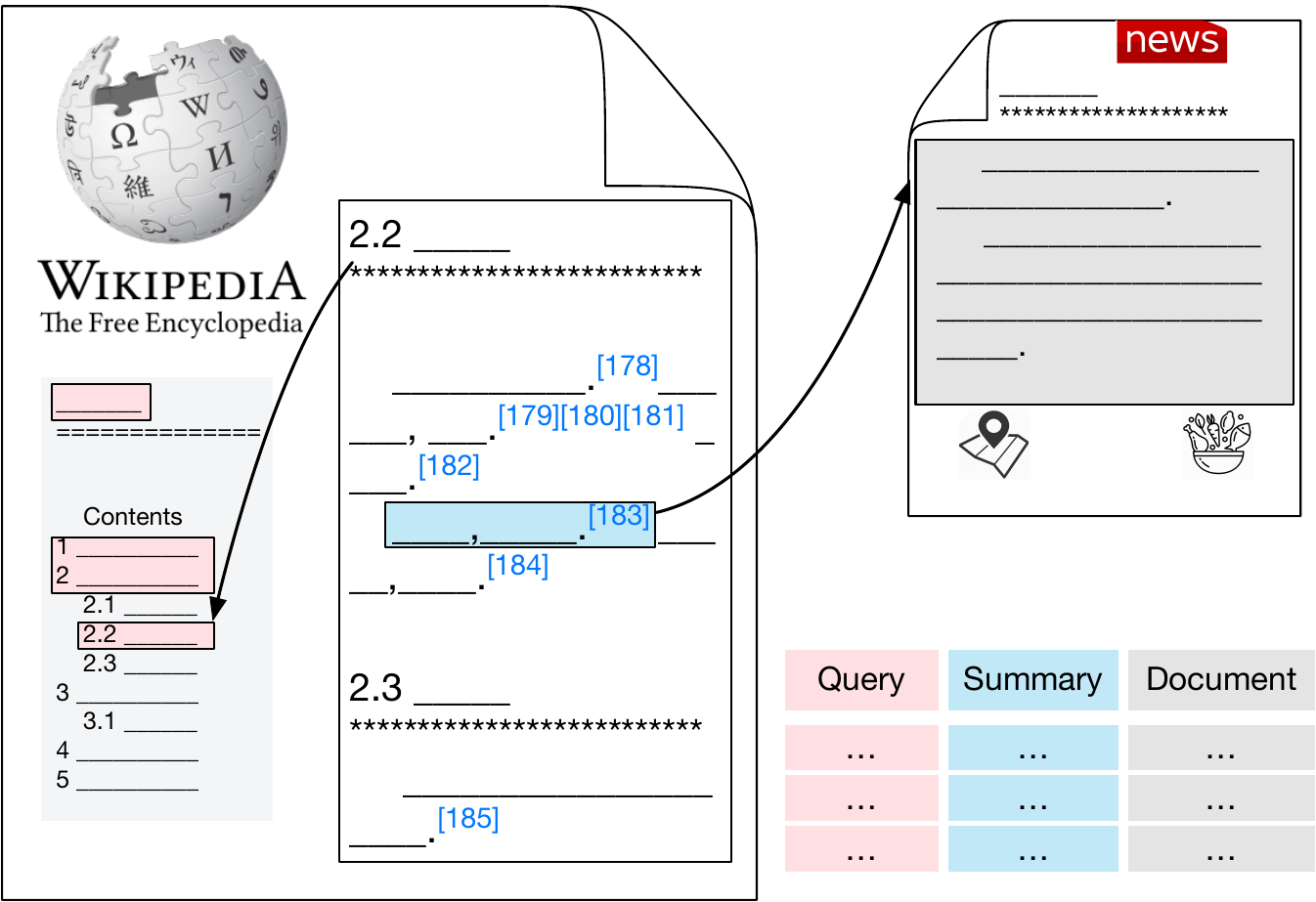}
    \caption{Illustration of \wikiref{} examples creation using \wiki{} and reference pages.}
    \label{fig:creation}
\end{figure}

\section{\wikiref{}: Transforming Wikipedia into Query-Focused Summarization Dataset}

We automatically construct a query-focused summarization dataset (named as \wikiref{}) using \wiki{} and corresponding reference web pages.
In the following sections, we will first elaborate on the creation process. Then we will analyze the queries, documents and summaries quantitatively and qualitatively.

\subsection{Data Creation}

We follow two steps to collect and process the data: (1) we crawl and parse English \wiki{} articles and their references to form raw examples; (2) we process and filter the raw examples through a set of fine-grained rules. 

\subsubsection{Raw Data Collection}
In the first step, we parse the English \wiki{} database dump into plain text and save statements with citations.
To maintain the highest standards possible, most statements in Wikipedia are attributed to reliable, published sources that can be accessed through hyperlinks.
If a statement is attributed to multiple citations, only the first citation is used.
We do not extend it to multi-document summarization.
We limit the sources of the citations to four types, namely web pages, newspaper articles, press and press releases.\footnote{Citation types including book, journal, AV media, Wikidata, album notes, comic, conference, court, act, encyclopedia, episode, mailing list, map, news group, patent, thesis and video game are skipped.}

A query-focused summarization example consists of a summary, a document and a query.
We find that the statement can be seen as a summary of the supporting citations from a certain aspect.
Therefore, we can take the title and the body of the citation\footnote{pypi.org/project/newspaper is used for downloading and parsing.} as the document and treat the statement as the summary.
We also find section titles are rough indicators of what aspects the statements focus on.
So we stack the article title and the section titles to form the query.
It is worth noticing that the queries are keywords, instead of natural language text as in other query-focused summarization datasets.
Now we have all the constituents a query-focused summarization example needs.

We illustrate the raw data collection process in Figure~\ref{fig:creation}.
The associated query, summary and the document are highlighted in colors in the diagram.
Eventually, we have collected more than $2,000,000$ examples through the raw data collection step.

\subsubsection{Data Curation}
To make sure the statement is a plausible summary of the cited document, we process and filter the examples through a set of fine-grained rules.
The texts are tokenized and lemmatized using Spacy\footnote{spacy.io}.

\begin{table}[!t]
    \centering
    \caption{Statistics of training set, development set and test set of the \wikiref{} dataset.}
    \begin{tabular}{lrrr}
    \toprule
         & Train & Dev & Test \\
         \midrule
          Total Examples & 256,724 & 12,000 & 12,000 \\
          Wiki Articles & 160,223 & 11,457 & 11,476 \\
          Document Tokens & 397.7 & 395.4 & 398.7 \\
          Document Sents & 18.8 & 18.7 & 18.8 \\
          Summary Tokens & 36.1 & 35.9 & 36.2 \\
          Summary Sents & 1.4 & 1.4 & 1.4 \\
          Query Depth & 2.5 & 2.5 & 2.5 \\
          Query Tokens & 6.7 & 6.8 & 6.7 \\
\bottomrule
    \end{tabular}
    \label{tab:dataset_statistics}
\end{table}

First, we calculate the unigram recall of the summary with reference to the document, where only the non-stop words are considered.
We throw out the example whose score is lower than the threshold.
Here we set the threshold to $0.5$ empirically, which means at least more than half of the summary tokens should be in the document.
It controls the quality of the dataset rather than restricts the summaries to be strictly extractive.

Next, we filter the examples under multiple length and sentence number constraints.
To set reasonable thresholds, we get the statistics of the examples survived in the previous step.
The $5$th and the $95$th percentiles are used as low and high thresholds of each constraint.

Finally, to make sure generating the summary with the given document is feasible, we filter the examples by extractive oracle score.
The extractive oracle is obtained through a greedy search over sentence combinations with no more than $5$ sentences.
\rouge{}-2 recall is the scoring metric and only the examples with an oracle score higher than $0.2$ are kept.

After running through the above curation steps, we have the \wikiref{} dataset with $280,724$ examples.
We randomly split the data into training, development and test sets, and ensure no overlapping documents across splits.

\subsection{Data Statistics}

\begin{table}[t]
    \centering
    \caption{Percentiles for different aspects of the whole  \wikiref{} dataset.}
    \begin{tabular}{lrrrrrrr}
    \toprule
         & 5 & 20 & 40 & 50 & 60 & 80 & 95 \\
         \midrule
          Document Tokens & 208& 267& 346& 387& 431& 530& 618 \\
          Document Sents & 9& 12& 16& 18& 20& 25& 33 \\
          Summary Tokens & 14& 20& 27& 31& 36& 50& 75 \\
          Summary Sents & 1& 1& 1& 1& 1& 2& 3 \\
          Query Depth & 2& 2& 2& 2& 3& 3& 4 \\
          Query Tokens & 3& 4& 5& 6& 7& 9& 13 \\
\bottomrule
    \end{tabular}
    \label{tab:dataset_percentile}
\end{table}

\begin{table}[!ht]
\caption{Quality rating results of human evaluation on the \wikiref{} dataset. 
``Relatedness'' indicates the relatedness of the summary and the query.
``Salience'' indicates to what extent the summary conveys the salient document content.
Two metrics are scored from 1 to 3, the higher the better.
}
\centering
\begin{tabular}{ccc}
\toprule
\textsc{Oracle Interval} & \tabincell{c}{\textsc{Relatedness}} & \tabincell{c}{\textsc{Salience}} \\
\midrule
20 $\sim$ 30 & 2.87 & 2.33 \\
30 $\sim$ 50 & 2.80 & 2.40 \\
50 $\sim$ 70 & 2.87 & 2.53 \\
70 $\sim$ 100 & 2.93 & 2.60 \\
\bottomrule
\end{tabular}
\label{tab:wikiref_huamn_eval}
\end{table}

Table~\ref{tab:dataset_statistics} and Table~\ref{tab:dataset_percentile} show statistics of the \wikiref{} dataset.
The development set and the test set contains $12,000$ examples each.
Statistics across splits are evenly distributed.
The numerous \wiki{} articles cover a wide range of topics.
The average depth of the query is $2.5$ with article titles are considered.
Since the queries are keywords in \wikiref{}, it is relatively shorter than the natural language queries with an average length of $6.7$ tokens.
Most summaries are composed of one or two sentences.
The document contains $18.8$ sentences on average.

\subsection{Human Evaluation}

We also conduct a human evaluation on $60$ \wikiref{} samples to examine the quality of the automatically constructed data.
We partition the examples into four bins according to the oracle scores and then sample $15$ examples from each bin.
Each example is scored by three volunteers in two criteria: (1) ``Relatedness'' examines to what extent the summary is a good response to the query and (2) ``Salience'' examines to what extent the summary conveys salient document content given the query.
Three participants are asked to score each example from 1 to 3.

Table~\ref{tab:wikiref_huamn_eval} shows the evaluation results.
We can see that the summaries are good responses to the queries across bins.
Since we take section titles as the query and the statement under the section as the summary, the high evaluation score can be attributed to high-quality \wiki{} pages.
When the oracle scores are getting higher, the summaries continue to better convey the salient document content specified by the query.
On the other hand, we notice that sometimes the summaries only contain a proportion of salient document content.
But it is acceptable to use for data augmentation purposes.

\section{Methodology}

\begin{figure*}[!t]
\centering
\includegraphics[width=0.95\textwidth]{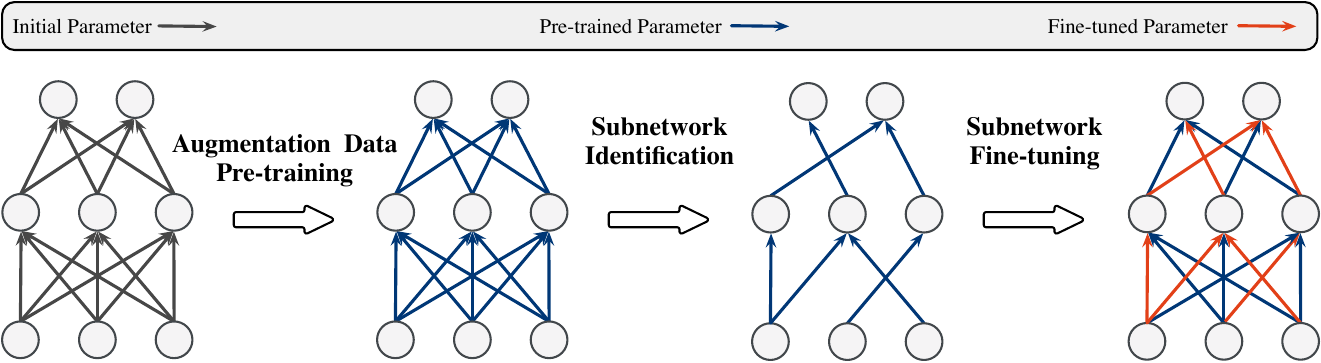}
\caption{The overview of our method based on data augmentation.
We first pre-train on the augmentation data, i.e., \wikiref{}. Then we identify a sparse subnetwork whose parameters are fine-tuned on the small benchmarks.}
\label{fig:model}
\end{figure*}

Figure~\ref{fig:model} gives an overview of our method based on data augmentation.
We first pre-train the model on the automatically constructed augmentation data, and then fine-tune model parameters on the human annotated benchmarks.
To train large models more efficiently on small benchmarks, we apply a subnetwork fine-tuning strategy (ST), which identifies sparse subnetworks in the model and only update the corresponding fraction of the parameters.
In the following, we will first describe our BERT-based query-focused summarization model (Q-BERT).
Then we introduce the sparse subnetwork fine-tuning strategy.

\subsection{Input Representation}

The query $\mathcal{Q}=(q_1, q_2,...,q_m)$ of $m$ tokens sequence and the document $\mathcal{D}=(s_1, s_2, ..., s_n)$ containing $n$ sentences are flattened and packed as a token sequence as input.
Following the standard practice of BERT, the input representation of each token is constructed by summing the corresponding token, segmentation and position embeddings.
Token embeddings projects the one-hot input tokens into dense vector representations.
Two segment embeddings $\mathbf{E}_Q$ and $\mathbf{E}_D$ are used to indicate query and document tokens respectively.
Position embeddings indicate the absolute position of each token in the input sequence.
To embody the hierarchical structure of the query in the sequential input, we insert a \textrm{[L\#]} token before the \textit{\#}-th query token sequence.
We also insert a \textrm{[CLS]} token at the beginning and a \textrm{[SEP]} token at the end. 

\subsection{BERT Encoding Layer}
In the encoding layer, we use BERT~\cite{bert}, a deep Transformer~\cite{transformer} consisting of stacked self-attention layers, as the encoder to aggregate query, intra-sentence and inter-sentence information into token-level encodings.
Given the packed input embeddings $\mathbf{H}^0=\left[\mathbf{x}_1,...,\mathbf{x}_{|x|}\right]$, we apply an $L$-layer Transformer to encode the input:
\begin{align}
    \mathbf{H}^l=\mathrm{Transformer}_l(\mathbf{H}^{l-1})
\end{align}
where $l\in\left[1,L\right]$.
The embedding of the $i$-th input token is $\mathbf{h}_i^L$.

\subsection{Query-Focused Pooling Layer}
\label{sec:qfs_pooling}
To get the query-focused sentence-level representation of each document sentence, we pool the token-level BERT encodings.
We first have the vector representation of the query via mean pooling:
\begin{align}
    \mathbf{h} _Q=\mathrm{MeanPool} ([\mathbf{h}^L_1,\dots , \mathbf{h}^L_m])
    \label{eq:query_pooling}
\end{align}
Then for the token encodings $\mathbf{H}_{j}^L\in\mathbb{R}^{d\times|s_j|}$ of the $j$-th sentence, we apply the weighted mean pooling to get its vector representation:
\begin{align}
\mathbf{v}_j=\mathrm{softmax}(\mathbf{h} _Q\mathbf{W}_q\mathbf{H}_{j}^L)\mathbf{H}_{j}^L
\label{eq:sent_pooling}
\end{align}
where $\mathbf{W}_q$ is a trainable parameter matrix.
The weight of each token in the sentence is determined according to the query.

\subsection{Output Layer}
The output layer is used to score sentences for extractive query-focused summarization.
Given $\mathbf{v}_j\in\mathbb{R}^d$ is the vector representation for the \textit{j}-th sentence.
When the model is supervised by sentence classification , the output layer is a linear layer followed by a \textrm{sigmoid} function $\sigma$:
\begin{align}
    P(s_j\vert\mathcal{Q},\mathcal{D})=\sigma(\mathbf{W}_c\mathbf{v}_j+\mathbf{b}_c)
\end{align}
where $\mathbf{W}_c$ and $\mathbf{b}_c$ are trainable parameters.
The output is the probability of including the \textit{i}-th sentence in the summary.

When it comes to sentence regression, a linear layer without activation function is used to estimate the score of a sentence:
\begin{align}
    \textrm{r}(s_j\vert\mathcal{Q},\mathcal{D})=\mathbf{W}_r\mathbf{v}_j+\mathbf{b}_r
\end{align}
where $\mathbf{W}_r$ and $\mathbf{b}_r$ are trainable parameters.

\subsection{Model Training and Inference}
The training objective of sentence classification is to minimize the binary cross-entropy loss:
\begin{equation}
\begin{aligned}
    \mathcal{L}=-\sum_i^n &y_i\textrm{log}P(s_i\vert\mathcal{Q},\mathcal{D})) + 
    (1-y_i)\textrm{log}(1-P(s_i\vert\mathcal{Q},\mathcal{D}))
\end{aligned}
\end{equation}
where
$y_i\in\{0,1\}$ is the oracle label of the \textit{i}-th sentence.

Training sentence regression model is to minimize the mean square error between the estimated score and the oracle score:
\begin{align}
    \mathcal{L}=\frac{1}{n}\sum_i^n (\textrm{r}(s_i\vert\mathcal{Q},\mathcal{D}) - \textrm{g}(s_i\vert\mathcal{S}_{ref}))^2
    \label{eq:mse}
\end{align}
where $\mathcal{S}_{ref}$ is the reference summary and
$\textrm{g}(s_i\vert\mathcal{S}_{ref})$ is the oracle score of the \textit{i}-th sentence.

During inference, a query-specific subset of $ \mathcal{D}$ is extracted as the output summary $\mathcal{\hat{S}}$, subject to a length constraint $l_c$:
\begin{equation}
\begin{aligned}
    &\hat{\mathcal{S}}=\underset{\mathcal{S} \subseteq \mathcal{D}}{\mathrm{argmax}} \sum_{s_i \in \mathcal{S}} \mathcal{M}(s_i\vert\mathcal{Q},\mathcal{D};\theta) \\
    &\mathrm{s.t.} \quad \sum_{s_i \in \mathcal{S}}\vert s_i \vert \le l_c
\end{aligned}
\end{equation}
where $\mathcal{M}$ is a sentence scoring model and $\theta$ are the parameters.

\subsection{Sparse Subnetwork Fine-tuning (ST)}
\label{sec:subnet_finding}
Our Q-BERT model contains tremendous number of parameters, first pre-trained on the augmentation data, and then fine-tuned on the small benchmarks.
The mismatch between the small data size and the massive number of parameters renders updating all parameters challenging.
Thus, we exploit a small fraction of parameters that deliberately selected for query-focused summarization by identifying sparse subnetworks in the pre-trained model.

We adopt a simple method to identify sparse subnetworks by including the largest magnitude parameters of the BERT encoding layer.
It is done in one step and does not require any annotated samples.
We compare the magnitude of parameters within each parameter matrix independently.
That is, the percentage of the remaining parameters in each parameter matrix is the same.
It avoids pruning some parameter matrices completely, especially with high sparsity.
The input embedding layer and output layer are always kept intact in subnetworks.
Note that all parameters participate in the forward computation, but only the parameters of the subnetworks are updated.

\section{Experiments on \wikiref{}}

In this section, we elaborate on experimental environment, model training and evaluation metrics.
We then present the results of benchmarking \wikiref{} as a standard query-focused summarization dataset.

\subsection{Implementation Details}

We use the uncased version of BERT-base for experiments.
The max sequence length is set to $512$.
We use Adam optimizer~\cite{kingma:adam} with learning rate of $3e-5$, $\beta_1$ = $0.9$, $\beta_2$ = $0.999$, L2 weight decay of $0.01$, and linear decay of the learning rate.
We split long documents into multiple windows with a stride of $100$.
Therefore, a sentence can appear in more than one window.
To avoid making predictions on an incomplete sentence or with a suboptimal context, we score a sentence only when it is completely included and its context is maximally covered.
We search for the best training epoch out of \{$3, 4$\} and select bath size out of \{$24, 32$\}.
We run all experiments with NVIDIA V100-32GB GPU cards using the PyTorch framework~\cite{pytorch} and the Hugging Face Transformer library~\cite{wolf-etal-2020-transformers}.

\begin{table*}[t]
\caption{\rouge{} scores  of baselines and the proposed model on \wikiref{} dataset.
}
\centering
\begin{tabular}{lccc c ccc}
\toprule
\multirow{2}{*}{Systems} & \multicolumn{3}{c}{Dev} && \multicolumn{3}{c}{Test} \\
\cmidrule{2-4} \cmidrule{6-8}
& ROUGE-1 & ROUGE-2 & ROUGE-L && ROUGE-1 & ROUGE-2 & ROUGE-L \\
\midrule
\textsc{All} & 14.05 & 7.84 & 12.97 && 14.09 & 7.88 & 13.01 \\
\textsc{Lead} & 26.55 & 10.66 & 21.99 && 26.32 & 10.48 & 21.81 \\ \midrule
\multicolumn{3}{l}{\textit{Regression Supvervision}} \\
\textsc{Bert} & 34.52 & 17.96 & 29.29 && 34.42 & 17.88 & 29.17  \\
\textsc{Q-Bert} & 34.83 & 18.18 & 29.51 && 34.58 &  17.92 & 29.27 \\
\hdashline
\textsc{Oracle} & 51.34 & 35.80 & 45.62 && 51.41 & 35.89 & 45.68 \\
\midrule
\multicolumn{3}{l}{\textit{Classification Supervision}} \\
\textsc{Transformer} & 28.18 & 12.92 & 23.92 && 28.07 & 12.80 & 23.79 \\ 
\textsc{Bert} & 35.40 & 18.49 & 30.30 && 35.08 & 18.15 & 29.99 \\
\textsc{Bert} w/o Query & 32.91 & 16.05 & 27.97 && 32.52 & 15.83 & 27.65 \\
\textsc{Q-Bert} & \textbf{35.85} & \textbf{18.91} & \textbf{30.59} && \textbf{35.50} & \textbf{18.45} & \textbf{30.37} \\
\hdashline
\textsc{Oracle} & 54.34 & 37.39 & 48.34 && 54.46 & 37.52 & 48.51 \\
\bottomrule
\end{tabular}
\label{tab:baselines}
\end{table*}

\subsection{Evaluation Metrics}
We use \rouge{}~\cite{lin-2004-rouge} as our automatic evaluation metric.
\rouge{}\footnote{ROUGE-1.5.5} is the official metric of the DUC benchmarks and widely used for summarization evaluation.
\rouge-N measures the summary quality by counting overlapping N-grams with respect to the reference summary.
\rouge{}-L measures the longest common subsequence.
\rouge{}-SU considers skip-gram with unigrams concurrence.

\subsection{Settings}
Training extractive summarization models to requires ground-truth labels for document sentences.
However, we can not find the sentences that exactly match the reference summary for most examples.
In order to solve the problem, we use a greedy algorithm similar to~\cite{zhou-etal-2018-neural-document} to find an oracle summary with document sentences that maximizes the \rouge{}-2 F1 score with respect to the reference summary.
Given a document of $n$ sentences, we greedily enumerate the combination of sentences.
For documents that contain numerous sentences, searching for a global optimal combination of sentences is computationally expensive.
Meanwhile, it is unnecessary since the reference summaries contain no more than four sentences.
So we stop searching when no combination with $i$ sentences scores higher than the best combination with $i$-1 sentences.
When training models with sentence regression supervision, the oracle score is its \rouge{}-2 F1 score.

During inference, we rank sentences according to their predicted scores.
Then we append the sentence one by one to form the summary if it scores higher than a threshold and is not redundant.
We skip the redundant sentences that contain overlapping trigrams concerning the current output summary as in~\cite{ft-bert-extractive}.
The threshold is searched on the development set to obtain the highest \rouge{}-2 F1 score.

\subsection{Baselines}
We compare the proposed model with the following baselines.
\paragraph{\textsc{All}} outputs all sentences of the document as a summary.
\paragraph{\textsc{Lead}} selects the leading sentences. We take the first two sentences for that the ground-truth summary contains $1.4$ sentences on average.
\paragraph{\textsc{BERT}} is a pre-trained language model based on Transformer. We append a [CLS] token to each sentence and use its token-level encoding to fine-tune BERT.
\paragraph{\textsc{Transformer}} uses the same structure as the BERT with randomly initialized weights.

\subsection{Results}

We report \rouge{}-1, \rouge{}-2 and \rouge{}-L scores\footnote{-n 2 -m -c 95 -r 1000}.
As shown in Table~\ref{tab:baselines}, Q-BERT with query-focused pooling layer outperforms all baselines and the strong BERT model.
The improvements are more pronounced when training with sentence classification supervision.
On average, the output summary consists of $1.8$ sentences.
\textsc{Lead} is a strong unsupervised baseline that achieves comparable results with the supervised neural baseline Transformer.
Even though \wikiref{} is a large-scale dataset, training models with parameters initialized from BERT still significantly outperform Transformer.
The model trained using sentence regression performs worse than the one supervised by sentence classification.
It is in accordance with oracle labels and scores as expected.
We observe a performance drop when generating summaries without queries (see ``w/o Query'').
It proves that the summaries in \wikiref{} are indeed query-focused.

\section{Experiments on DUC Benchmarks}

\subsection{Dataset}
The documents of DUC are from the news domain and grouped into clusters according to their topics.
The summary is required to be no longer than $250$ tokens.
Table~\ref{tab:duc_stat} shows statistics of the DUC datasets.
Each document cluster has
several reference summaries generated by humans and 
a query that specifies the focused aspects and desired information.
We show an example query from the DUC 2006 dataset below:
\begin{quote}
\small
EgyptAir Flight 990? \\
What caused the crash of EgyptAir Flight 990? \\
Include evidence, theories and speculation.
\end{quote}
The first narrative is usually a title and followed by several natural language questions or narratives. 

\subsection{Settings}
We follow the standard practice to alternately train our model on two years of data and test on the third.
The oracle scores used in model training are \rouge{}-2 recall of sentences.
In this paper, we score a sentence by only considering the query and its document.
Then we rank sentences according to the estimated scores across documents within a cluster.
For each cluster, we fetch the top-ranked sentences iteratively into the output summary with redundancy constraints met.
A sentence is redundant if more than half of its bigrams appear in the current output summary.
To be comparable with previous work on DUC datasets, we report the \rouge{}-1 and \rouge{}-2 recall computed with official parameters~\footnote{-n 2 -x -m -2 4 -u -c 95 -r 1000 -f A -p 0.5 -t 0 -l 250} that limits the length to 250 words.

The \wikiref{} dataset is used as augmentation data for DUC datasets in two steps.
We first pre-train a summarization model on the \wikiref{} dataset.
Subsequently, we use the DUC datasets to further fine-tune parameters of the best pre-trained model.
We use 10\% parameters for sparse subnetwork fine-tuning due to its optimal performance.

\begin{table}[t]
\caption{Statistics of DUC benchmarks.}
\centering
\begin{tabular}{lccc}
\toprule
& 2005 & 2006 & 2007 \\
\midrule
Clusters & 50 & 50 & 45 \\
Documents & 1,593 & 1,250 & 1,125 \\
Sentences & 46,033 &34,585 & 24,176 \\
\bottomrule
\end{tabular}
\label{tab:duc_stat}
\end{table}

\subsection{Baselines}

We compare our method with several previous query-focused summarization models:

\paragraph{\textsc{Lead}}
is a simple baseline that selects the leading sentences to form a summary.

\paragraph{\textsc{Query-Sim}}
is an unsupervised method that ranks sentences according to its TF-IDF cosine similarity to the query.

\paragraph{\textsc{Svr}~\cite{svr}}
is a supervised baseline that extracts both query-dependent and query-independent features and then using Support Vector Regression to learn the weights of features.

\paragraph{\textsc{AttSum}~\cite{cao-attsum}}
is a neural attention summarization system that tackles query relevance ranking and sentence salience ranking jointly.

\paragraph{\textsc{CrSum}~\cite{Ren-crsum} }
is the contextual relation-based neural summarization system that improves sentence scoring by utilizing contextual relations among sentences.

\paragraph{\textsc{PQSUM}~\cite{wsl-dl}} uses the BERTSUM~\cite{ft-bert-extractive} model pre-trained on the CNN/DailyMail dataset~\cite{cnndaily}.

\paragraph{\textsc{QuerySum}~\cite{xu-lapata-2020-coarse}} is a coarse-to-fine framework with its query relevance estimator trained with external question answering datasets.

\subsection{Results}
\begin{table*}[t]
\caption{\rouge{} scores on the DUC 2005, 2006 and 2007 benchmarks.
``DA'' is short for data augmentation using the \wikiref{} dataset.
``DA Pre-trained'' denotes applying the model pre-trained on augmentation data to DUC without fine-tuning.
``ST'' is short for subnetwork fine-tuning.
}
\centering 
\begin{tabular}[width=\textwidth]{lcc c cc c cc}
\toprule
 \multirow{2}{*}{Systems} & \multicolumn{2}{c}{DUC 2005} && \multicolumn{2}{c}{DUC 2006} && \multicolumn{2}{c}{DUC 2007}\\
 \cmidrule{2-3} \cmidrule{5-6} \cmidrule{8-9}
 & ROUGE-1 & ROUGE-2 && ROUGE-1 & ROUGE-2 && ROUGE-1 & ROUGE-2 \\
 \midrule
 \multicolumn{3}{l}{\textit{Without Data Augmentation}} \\
 \textsc{Lead}~\cite{Ren-crsum} & 29.71 & 4.69 && 32.61 & 5.71 && 36.14 & 8.12 \\
 \textsc{Query-Sim}~\cite{Ren-crsum} & 32.95 & 5.91 && 35.52 & 7.10 && 36.32 & 7.94 \\
 \textsc{Svr}~\cite{svr} & 36.91 & 7.04 && 39.24 & 8.87 && 43.42 & 11.10 \\
 \textsc{CrSum}~\cite{Ren-crsum} & 36.96 & 7.01 && 39.51 & 9.19 && 41.20 & 11.17 \\
 \textsc{AttSum}~\cite{cao-attsum} & 37.01 & 6.99 && 40.90 & 9.40 && \textbf{43.92} & 11.55 \\
 \textsc{QuerySum}~\cite{xu-lapata-2020-coarse} & - & - && \textbf{41.60} & 9.50 && 43.30 & \textbf{11.60} \\
 \textsc{Bert} & \textbf{37.82} & \textbf{7.88} && 41.35 & \textbf{9.60} && 43.55 & 11.39 \\
 \textsc{Q-Bert} & 37.47 & 7.58 && 40.88 & 9.54 && 42.84 & 11.24 \\
 \midrule
 \multicolumn{3}{l}{\textit{Classification Pre-training}} \\
 \textsc{PQSUM}~\cite{wsl-dl} & 37.55 & 7.84 && 40.41 & 9.22 && 42.41 & 11.08  \\
 \textsc{DA Pre-trained} & 36.19 & 7.00 && 38.67 & 7.88 && 40.08 & 9.19 \\
 \textsc{Bert + DA} & 38.77 & 8.31 && 41.65 & 10.04 && 44.31 & 11.85 \\
 \midrule
 
 \multicolumn{3}{l}{\textit{Regression Pre-training}} \\
 \textsc{DA Pre-trained} & 36.52 & 7.02 && 38.81 & 8.37 && 41.09 & 10.29 \\
 \textsc{Bert + DA} & 38.44 & 8.33 && 41.64 & 9.98 && 44.14 & 12.12 \\
 \textsc{Q-Bert + DA} & 38.68 & 8.51 && 41.81 & 10.17 && 44.72 & 12.43 \\
 \textsc{Q-Bert + DA + ST} & \textbf{39.21} & \textbf{8.62} && \textbf{41.96} & \textbf{10.29} && \textbf{45.06} & \textbf{12.55} \\
 \hdashline
 \textsc{Oracle} & 43.71 & 13.77 && 48.02 & 17.22 && 49.80 & 19.19 \\
 \bottomrule
\end{tabular}
\label{tab:duc_results}
\end{table*}

Table~\ref{tab:duc_results} shows the \rouge{} scores of previous models and our proposed method.
Fine-tuning BERT on DUC datasets alone obtains comparable results.
Our data augmentation method advances the model to a higher performance on all DUC benchmarks.
And the Q-BERT and the sparse subnetwork fine-tuning both further improve the results.
We also notice that models pre-trained on the augmentation data achieve reasonable performance without further fine-tuning model parameters.
It implies the \wikiref{} dataset reveals useful knowledge shared by the DUC dataset.
We pre-train models on augmentation data under both sentence classification and sentence regression supervision.
Since the training objectives of pre-training and fine-tuning are the same, the performance of pre-training with regression supervision is slightly better.

Without data augmentation, fine-tuning the more complex Q-BERT performs worse than BERT on the tiny DUC benchmarks.
However, by pre-training on the augmentation data, the advantages of Q-BERT are revealed and better results are achieved.
It shows that the small size of DUC benchmarks renders training data-driven neural models difficult and hinders the development of more effective architectures for query-focused summarization.

\subsection{Human Evaluation}

\begin{table}[t]
\caption{Human evaluation results on the DUC 2007 benchmark.
Two metrics are scored from 1 to 3, the higher the better.
}
\centering
\begin{tabular}{lcc}
\toprule 
 & \tabincell{c}{\textsc{Relatedness}} & \tabincell{c}{\textsc{Redundancy}} \\
\midrule
\textsc{Bert} & 2.20 & 2.75 \\
\textsc{Bert + DA} & 2.48 & 2.78 \\
\bottomrule
\end{tabular}
\label{tab:duc_huamn_eval}
\end{table}

We conduct a human evaluation of the output summaries before and after applying data augmentation.
We sample $30$ examples from the DUC 2007 dataset for analysis.
Three volunteers are asked to score the outputs on a $1$-$3$ scale.
The results are shown in Table~\ref{tab:duc_huamn_eval}.
We can see that the model augmented by the \wikiref{} dataset produces more query-related summaries.
We attribute this to the improved coverage and query-focused extraction brought by the large-scale augmentation data. 
As to the redundancy, the \wikiref{} dataset yields no significant effect to produce less redundant summaries.

\section{Analyses and Discussion}

\begin{figure}[!t]
\centering
\includegraphics[width=\linewidth]{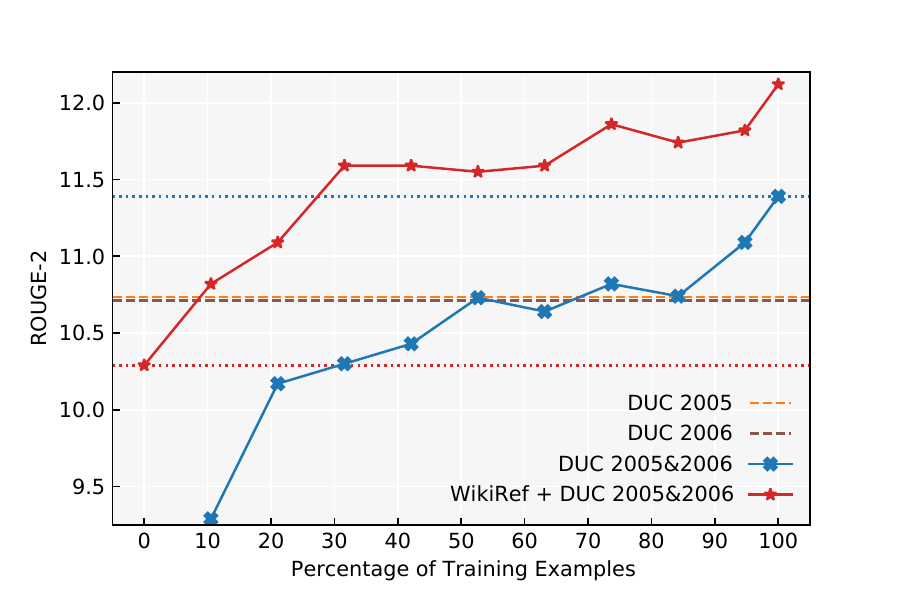}
\caption{
\rouge{}-2 score on the DUC 2007 with various number of training data.
The x-axis shows the percentage of used training data.
The horizontal lines indicate using DUC 2005 or DUC 2006 for training.
The red line indicates using \wikiref{} as augmentation data for pre-training and the blue line does not.
}
\label{fig:aug_ablation}
\end{figure}
\subsection{The Effectiveness of Data Augmentation}
To further analysis the effectiveness of our data augmentation method, we vary the number of human annotated examples for fine-tuning BERT or model pre-trained on \wikiref{}.
Here we take DUC 2007 for evaluation, and the joint of DUC 2005 and 2006 for training.
As shown in Figure~\ref{fig:aug_ablation}, we can see that our data augmentation method obtains consistent improvement over the BERT fine-tuning with various number of data.
Using either DUC 2005 alone or DUC 2006 alone yields inferior performance than using both.
In addition, the \textit{red} horizontal dashed line shows that the pre-trained model performs as well as fine-tuning BERT using approximately 30\% of the data.
The \textit{blue} horizontal dashed line shows that using the same amount of the data with our data augmentation outperforms full data BERT fine-tuning.

Three characteristics of the \wikiref{} make it an effective augmentation data for the DUC benchmarks.
At first, their documents share the same domain.
The DUC documents are news articles.
We also crawl newspaper webpages as one source of the \wikiref{} documents.
Secondly, queries in the \wikiref{} dataset are in nature hierarchical that specify the focused aspects gradually.
This intrinsic is in line with the DUC queries composed of several narratives to specify the desired information.
It makes the model transfer to the DUC datasets more easily.
At last, we construct the \wikiref{} dataset to be a large-scale query-focused summarization dataset that contains more than $280,000$ examples, which is in sharp contrast to the DUC datasets containing only $145$ clusters with around $10,000$ documents in total.
The large size improves the coverage. 
Query relevance and sentence context can be better modeled using data-driven neural methods with \wikiref{}.
The above characteristics together contribute to the effective augmented data for query-focused summarization.

\begin{figure}[t]
\centering
\includegraphics[width=\linewidth]{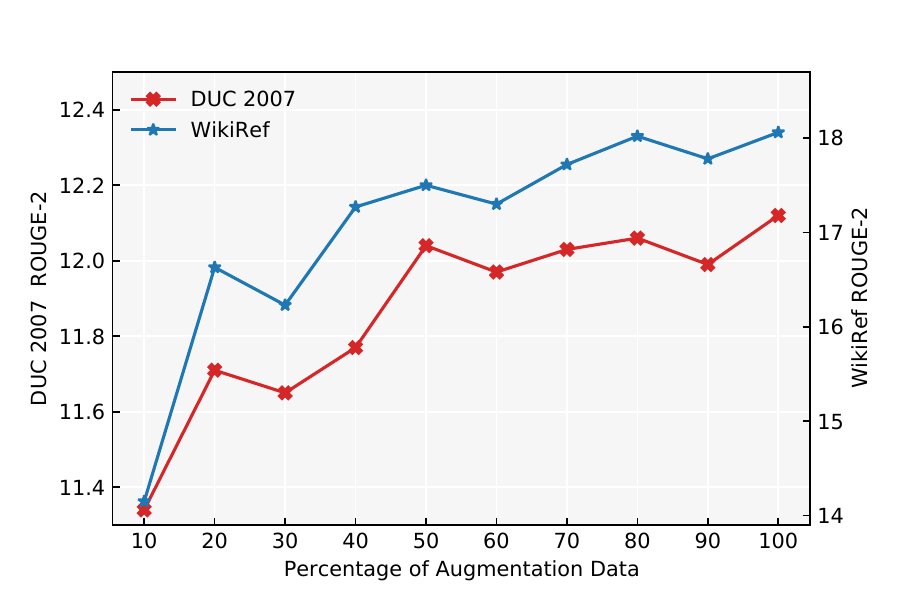}
\caption{
Trends of \rouge{}-2 scores with various number of \wikiref{} training data, which used as augmentation data for DUC 2007.
}
\label{fig:quantity}
\end{figure}

\begin{figure}[t]
\centering
\includegraphics[width=\linewidth]{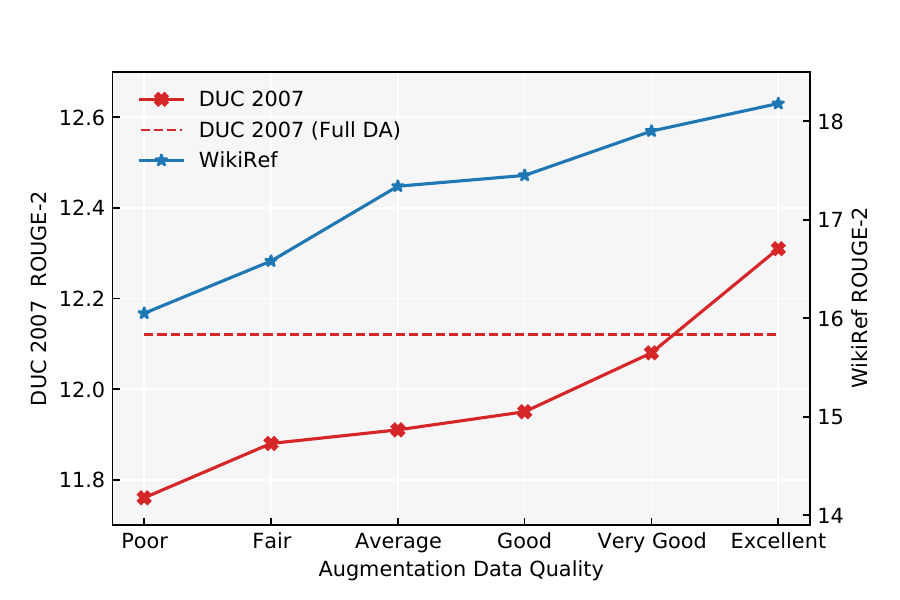}
\caption{
Trends of \rouge{}-2 scores with half of \wikiref{} training data of different quality.
The red horizontal dashed line indicates using the full \wikiref{} training set as augmentation data for DUC 2007.
}
\label{fig:quality}
\end{figure}

\subsection{Impact of Augmentation Data Size and Quality}
Firstly, we investigate the impact of data size on the effectiveness of data augmentation methods.
Figure~\ref{fig:quantity} shows that increasing the data size improves the performance on \wikiref{}.
However, the improvements on DUC 2007 seem to be saturated when using more than half of the augmentation data.
It could be partly due to that increasing the data size also introduces more noise.

Next, we analysis the impact of data quality.
The oracle \rouge{}-2 score is used as an proxy for  sample quality.
Specifically, we sort all the samples in ascending order and take half of the data with a step of 10\%.
In this way, we have six overlapped slices of data of increasing data quality, namely \textit{Poor, Fair, Average, Good, Very Good, Excellent}.
As shown in Figure~\ref{fig:quality}, higher data quality always yield better performance on both \wikiref{} and DUC 2007.
Furthermore, the highest quality slice with only half the data size outperforms the full augmentation data.
It shows that data quality is more important than data size and reducing the noise in the augmentation data can lead to further improvement.

\subsection{Effect of Pooling Methods}
\begin{table}[!t]
    \caption{Effectiveness of using different pooling methods to get sentence-level representations.}
    \centering
    \begin{tabular}{lcc c cc}
    \toprule
    \multirow{2}{*}{Model} & \multicolumn{2}{c}{\wikiref{}} && \multicolumn{2}{c}{DUC 2007} \\
    \cmidrule{2-3} \cmidrule{5-6}
     &  R-2 &  R-L &&  R-2 & R-SU4 \\
     \midrule
    \textsc{Bert} & 18.15 & 29.99 && 12.12 & 17.58 \\
    \textsc{\quad +Max Pooling} & 17.80 & 29.00 && 11.85  & 17.37 \\
    \textsc{\quad +Mean Pooling} & 18.26 & 29.93 && 11.93 & 17.47 \\
    \textsc{\quad +Attention Pooling} & 18.17 & 29.75 && 12.21 & 17.65  \\
    \textsc{Q-Bert} & 18.45 & 30.37 && 12.43 & 17.75 \\
    \bottomrule
    \end{tabular}
    \label{tab:model_ablation}
\end{table}

In Section~\ref{sec:qfs_pooling}, we apply query-focused pooling (Equation \ref{eq:query_pooling} to \ref{eq:sent_pooling}) to token-level encodings to get sentence vector.
We have tried several other alternatives to the query-focused pooling.
A special token is appended to each sentence and its token-level representation is instead used for prediction in BERT.
Attention pooling~\cite{wang-etal-2017-gated} is similar to the proposed query-focused pooling, except it uses a trainable vector independent of the input query to assign the weight of sentence tokens in Equation~\ref{eq:sent_pooling}.
As shown in Table~\ref{tab:model_ablation}, max pooling encodes the least information and performs the worst.
Mean pooling and attention pooling are comparable to BERT.
At last, our model with query-focused pooling that weighs sentence tokens adaptively according to the input query works best.

\subsection{Subnetwork Identification Methods}
\begin{table}[!t]
    \centering
    \caption{Performance of different subnetwork identification methods.}
    \begin{tabular}{llccc}
    \toprule
    Method & Init & R-1 & R-2 & R-SU4 \\
    \midrule
    None & \textsc{Bert} &  44.72 &	12.43 &	17.75  \\
    \textsc{Iterative} & \textsc{Bert} & 41.98 & 10.93	& 16.11 \\
    \textsc{Iterative} & \textsc{DA Pre-trained} & 44.54 & 12.28 & 17.58 \\
    \textsc{One-Step} & \textsc{DA Pre-trained} & 45.06 &	12.55 &	17.86  \\
    \bottomrule
    \end{tabular}
    \label{tab:prune_method}
\end{table}

In Section~\ref{sec:subnet_finding}, we identify the subnetwork in one step by keeping the largest magnitude parameters of the model pre-trained on augmentation data.
We have explored an iterative method~\cite{Zhu2018To} to identify subnetworks in models initialized by BERT or pre-trained on augmentation data.
This iteravtive method gradually reduces the size of subnetwork during training with augmentation data.
In Table~\ref{tab:prune_method}, we show the results of fine-tuning the identified subnetworks on DUC 2007 benchmark.
Fine-tuning subnetworks generated by the iterative method performs worse than that of the full model under the two initializations. 
The one-step method applied to the pre-trained model can yield subnetworks specific to the query-focused summarization task works best in our work.
Note that we do not apply one-step method to BERT because it generates the same general subnetwork for all tasks~\cite{DBLP:conf/emnlp/PrasannaRR20,lottery-for-bert,liang2021super}, which is not our focus.

\section{Related Work}
\label{section2}
A wide range of unsupervised approaches has been proposed for extractive summarization.
Surface features, such as n-gram overlapping, term frequency, document frequency, sentence positions~\cite{Ren-crsum}, sentence length~\cite{cao-attsum}, and TF-IDF cosine similarity~\cite{qfs_graph}.
Maximum Marginal Relevance (MMR)~\cite{mmr} greedily selects sentences and considered the trade-off between saliency and redundancy.
McDonald~\cite{ilp} treat sentence selection as an optimization problem and solve it using Integer Linear Programming (ILP).
Lin and Bilmes~\cite{lin2010multi} propose using submodular functions to maximize an objective function that considers the trade-off between coverage and redundancy terms.

Graph-based models make use of various inter-sentence and query-sentence relationships are also widely applied in the extractive summarization area.
LexRank~\cite{erkan2004lexrank} scores sentences in a graph of sentence similarities.
Wan and Xiao~\cite{qfs_graph} apply manifold ranking to make use of the sentence-to-sentence and sentence-to-document relationships and the sentence-to-query relationships.
We also model the above mentioned relationships, except for the cross-document relationships, like a graph at token level, which are aggregated into distributed representations of sentences.

Supervised methods with machine learning techniques~\cite{crf,svr,logist_regression} are also used to better estimate sentence importance. 
In recent years, few deep neural networks based approaches have been used for extractive document summarization.
Cao~\et~\cite{cao-attsum} propose an attention-base model that jointly handles sentence salience ranking and query relevance ranking.
It automatically generates distributed representations for sentences as well as the document.
To leverage contextual relations for sentence modeling, Ren~\et~\cite{Ren-crsum} propose CRSum that learns sentence representations and context representations jointly with a two-level attention mechanism.
Xu and Lapata~\cite{xu-lapata-2020-coarse} propose a coarse-to-fine framework which progressively estimates sentence relevance.
Kulkarni~\et~\cite{aquamuse} propose SIBERT that extends HIBERT~\cite{hibert} to query-focused multi-document summarization by introducing a cross-document infusion layer and incorporating queries as additional contexts.
The small data size is the main obstacle to develop neural models for query-focused summarization.

\wiki{} Wikipedia provides rich resources for exploring various natural language processing tasks. 
For text summarization, Liu~\et~\cite{j.2018generating} build WikiSum, a multi-document summarization dataset that uses both Wikipedia references and web search results to generate long Wikipedia article abstractively.
Base on WikiSum, Hayashi~\et~\cite{wikiasp} propose WikiAsp for aspect-based summarization, a subset of section titles is selected as the set of aspects for each domain.
Question answering datasets~\cite{kwiatkowski-etal-2019-natural} are also explored to mine query-based multi-document summarization examples, the ComSum dataset~\cite{aquamuse}, or train evidence estimator to improve the relatedness between the summary and the query~\cite{xu-lapata-2020-coarse}.
We use \wiki{} to automatically construct a large dataset for single document query-focused summarization.
It can be used as a standard dataset or as a means of data augmentation for small benchmarks.

\section{Conclusions}
In this paper, we propose to use \wiki{} articles and the corresponding references to automatically construct a large-scale query-focused summarization dataset named \wikiref{}.
The statements, supporting citations and article title along with section titles are used as summaries, documents and queries respectively.
The \wikiref{} dataset serves as a means of data augmentation for DUC benchmarks.
It is also shown to be an eligible query-focused summarization benchmark.
Moreover, we develop a BERT-based extractive query-focused summarization model and a sparse subnetwork fine-tuning method to improve the performance on tiny benchmarks.
The results on DUC benchmarks show that our augmentation data facilitate query-focused summarization and helps to develop more efficient data-driven models.
Quantitatively and qualitatively analysis shows that improving the augmentation data quality is more important than expanding the scale, which is promising in future work.


%

%



\ifCLASSOPTIONcaptionsoff
  \newpage
\fi



%
%
%

\bibliographystyle{IEEEtran}
\bibliography{IEEEtran}
%


\end{CJK*}
\end{document}